# *When will the mist clear?* On the Interpretability of Machine Learning for Medical Applications: a survey.

**Authors**

Antonio-Jesús Banegas-Luna[*, a] (ajbanegas@ucam.edu), Jorge Peña-García[a] (jpena@ucam.edu), Adrian Iftene[b] (adiftene@info.uaic.ro), Fiorella Guadagni[c,d] (guadagnifiorella@gmail.com), Patrizia Ferroni[c,d] (ferronipatrizia@gmail.com), Noemi Scarpato[d] (noemi.scarpato@uniroma5.it), Fabio Massimo Zanzotto[e] (fabio.massimo.zanzotto@uniroma2.it), Andrés Bueno-Crespo[a] (abueno@ucam.edu), Horacio Pérez-Sánchez[*, a] (hperez@ucam.edu)

**Affiliations**

[a] *Universidad Católica de Murcia (UCAM), Structural Bioinformatics and High-Performance Computing Research Group (BIO-HPC), Murcia, Spain.*
[b] *Universitatea Alexandru Ioan Cuza (UAIC), Faculty of Computer Science, Iaşi, Romania*
[c] IRCCS San Raffaele Pisana, *Rome, Italy*
[d] San Raffaele Roma Open University, Department of Human sciences and Promotion of the quality of life, Rome, Italy
[e] *University or Rome Tor Vergata, Dipartimento di Ingegneria dell'Impresa "Mario Lucertini", Rome, Italy*

**ORCID**

Antonio-Jesús Banegas-Luna: 0000-0003-1158-8877
Jorge Peña-García: 0000-0002-3622-4634
Adrian Iftene: 0000-0003-3564-8440
Fiorella Guadagni: 0000-0003-3652-0457
Patrizia Ferroni: 0000-0002-9877-8712
Noemi Scarpato: 0000-0002-6573-8095
Fabio Massimo Zanzotto: 0000-0002-7301-3596
Andrés Bueno-Crespo: 0000-0003-1734-6852
Horacio Pérez-Sánchez: 0000-0003-4468-7898

**Corresponding author***:* ajbanegas@ucam.edu (0034-968278819), hperez@ucam.edu (0034-968278819)

## Abstract

Artificial Intelligence is providing astonishing results, with medicine being one of its favourite playgrounds. In a few decades, computers may be capable of formulating diagnoses and choosing the correct treatment, while robots may perform surgical operations, and conversational agents could interact with patients as virtual coaches. Machine Learning and, in particular, Deep Neural Networks are behind this revolution. In this scenario, important decisions will be controlled by standalone machines that have learned predictive models from provided data. Among the most challenging targets of interest in medicine are cancer diagnosis and therapies but, to start this revolution, software tools need to be adapted to cover the new requirements. In this sense, learning tools are becoming a commodity in *Python* and *Matlab* libraries, just to name two, but to exploit all their possibilities, it is essential to fully understand how models are interpreted and which models are more interpretable than others.

In this survey, we analyse current machine learning models, frameworks, databases and other related tools as applied to medicine - specifically, to cancer research - and we discuss their interpretability, performance and the necessary input data. From the evidence available, ANN, LR and SVM have been observed to be the preferred models. Besides, CNNs, supported by the rapid development of GPUs and tensor-oriented programming libraries, are gaining in importance. However, the interpretability of results by doctors is rarely considered which is a factor that needs to be improved. We therefore consider this study to be a timely contribution to the issue.

## Abbreviations

1CM, One-carbo metabolism; AI, Artificial Intelligence; ANN, Artificial Neural Network; AUC, Area Under the Curve; BC, Breast Cancer; BioBIM, InterInstitutional Multidisciplinary Biobank; BMI, Body Mass Index; BN, Bayesian Network; CCF, Cancer Cell Fraction; CNN, Convolutional Neural Network; CRC, Colorectal Cancer; DCNN, Dilated Convolutional Neural Network; DL, Deep Learning; DSS, Decision Support System; DT, Decision Tree; ELM, Extreme Learning Machine; EMR, Electronic Medical Record; ENLR, Elastic Net Logistic Regression; FOLFIRI, 5-FU, leucovorin and irinotecan; FOLFOX, 5-FU, leucovorin and oxaliplatin; FT, Fourier Transform, GBM, Gradient Boosting Machine; GEO, Gene Expression Omnibus; GOSS, Genetic Ontology Similarity Score; GPU, Graphics Processing Unit; HDF5, Hierarchical Data Format 5; HNSCC, Head and Neck Squamous Cell Carcinoma; HPC, High Performance Computing; ICBC, Iranian Centre for Breast Cancer; IMRT, Intensity Modulated Radiotherapy; KNN, K-Nearest Neighbours; LDA, Linear Discriminant Analysis; LPP, Locality Preserving Projection; LR, Logistic Regression; LSTM, Long Short-Term Memory; ML, Machine Learning; MVA, Multivariate analysis; NCBI, National Center for Biotechnology Information; NCSS, Number Cruncher Statistical Systems; NMSC, Non-Melanoma Skin Cancer; PCA, Principal Component Analysis; RECIST, Response Evaluation Criteria In Solid Tumors; REVOLVER, Repeated EVOLution in cancER; RF, Random Forest; RMSE, Root Mean Square Error; RNN, Recurrent Neural Network; RO, Random Optimization; ROC, Receiver Operating Characteristic; SAP, Single Amino Acid Polymorphism; SEABED, Segmentation and Biomarker Enrichment of Differential Treatment Response; SEER, Surveillance, Epidemiology and End Results; SIFT, Sorting Intolerant From Tolerant; SKCM, Skin Cutaneous Melanoma; SNP, Single Nucleotide Polymorphism; SSL, Semi-Supervised Learning; SVC-W, Support Vector Classification with Weight; SVM, Support Vector Machine; SVM-L1, Support Vector Machine with L1 Regularization; TCGA, The Cancer Genome Atlas; TGF-β, Transforming Growth Factor beta; TL, Transfer Learning; WEKA-FCBF, Waikato Environment of Knowledge Analysis - Fast Correlation Based Filter; WHO, World Health Organization; XAI, Explainable Artificial Intelligence; YARN, Yet Another Resource Negotiator.

## 1. Introduction

Cancer has become one of the most common human diseases and causes of death (Cronin et al., 2018; Culp, Soerjomataram, Efstathiou, Bray, & Jemal, 2020; Ferlay et al., 2015). Among other factors, its occurrence is mainly growing because of aging (Chiavenna, Jaworski, & Vendrell, 2017). Even though cancer is a disease that affects men as well as women, there seems to be a clear relationship between gender and incidence. Thus, lung, prostate, colorectal, stomach and liver cancer are predominant among men, while breast, colorectal, lung, cervical and thyroid are the most common cancers in women (https://www.who.int/health-topics/cancer). Figure 1 depicts the number of estimated deaths in 2020 by cancer type collected from the Surveillance, Epidemiology and End Results (SEER) database.

| | | |
|---|---|---|
| Lung & bronchus | 63 220 | 22.15% |
| Breast | 42 170 | 14.78% |
| Colon & rectum | 24 570 | 8.61% |
| Pancreas | 22 410 | 7.85% |
| Ovary | 13 940 | 4.89% |
| Endometrium | 12 590 | 4.41% |
| Liver & intrahepatic bile duct | 10 140 | 3.55% |
| Leukemia | 9 680 | 3.39% |
| Non-Hodkin lymphoma | 8 480 | 2.97% |
| Brain & other nervous system | 7 830 | 2.74% |
| **All sites** | **285 360** | **100%** |

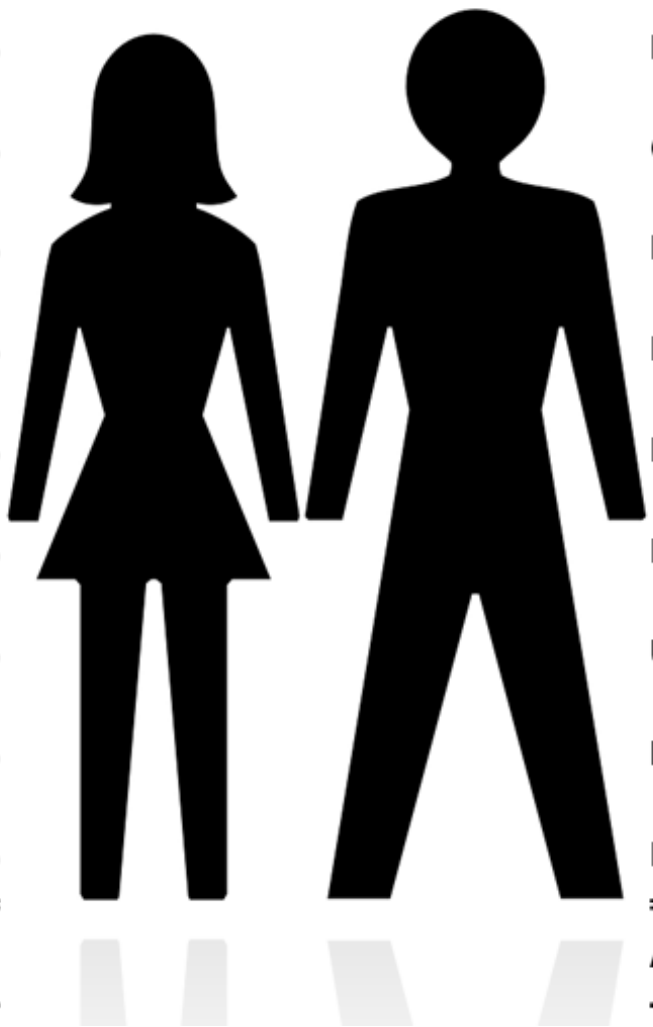

| | | |
|---|---|---|
| Lung & bronchus | 72 500 | 22.57% |
| Prostate | 33 330 | 10.38% |
| Colon & rectum | 28 630 | 8.91% |
| Pancreas | 24 640 | 7.67% |
| Liver & intrahepatic bile duct | 20 020 | 6.23% |
| Leukemia | 13 420 | 4.18% |
| Esophagus | 13 100 | 4.08% |
| Urinary bladder | 13 050 | 4.06% |
| Non-Hodkin lymphoma | 11 460 | 3.57% |
| Brain & other nervous system | 10 190 | 3.17% |
| **All sites** | **321 160** | **100%** |

**Fig. 1.** Estimated deaths in USA in 2020 by cancer type and gender. Source: SEER database.

A diverse range of therapies, including chemotherapy, radiotherapy, surgery and irradiation, is used in cancer patients depending on tumour type and stage. Unfortunately, the success of these treatments is limited because they attack normal and tumoral cells equally, which may result in toxicity and make the tumoral cells drug-resistant. In this scenario, early detection is a crucial factor for the successful application of therapies, for limiting associated side effects and, consequently, increasing the chance of survival (Coleman, 2017; Loud & Murphy, 2017). For this reason, providing the physicians with appropriate tools for accurate diagnosis and prognosis remains a major challenge in cancer research.

Colorectal cancer (CRC) is the third most common type of cancer worldwide, representing 10% of all diagnosed cases, and the fourth in the number of deaths it causes (Araghi et al., 2019; Dekker, Tanis, Vleugels, Kasi, & Wallace, 2019). Furthermore, these figures are not very promising because the number of CRC cases is expected to increase by around 60% in the forthcoming decade (Arnold et al., 2017).

As regard the reasons for such disheartening data, bad dietary habits are suspected to be behind the growing number of CRC cases reported in recent years but there are other reasons, such as the lack of exercise, obesity and smoking that are suspected of causing tumours (Kuipers et al., 2015). Moreover, familial and hereditary antecedents have proved to influence the incidence of this cancer (Weinberg, Marshall, & Salem, 2017). In an attempt to identify reasons, beyond the biological, for the evolution of CRC worldwide, Arnold (Arnold et al., 2017) published a study correlating the human development index with the incidence and high mortality of CRC, which resulted in the classification of countries into three groups with well-defined characteristics. In short, a number of factors in our daily lives promote the emergence of colorectal tumours and, although there is no clear numerical estimation of how much these factors contribute to the appearance CRC, it seems to be in our hands to change the trend. From a more medical point of view, the high morbidity and mortality rates could be explained by the fact that malignant CRC tumours are considered to be especially complex biologically (García-Figueiras et al., 2018).

Much effort has been put into predicting CRC or, at least, into predicting the manner in which the tumour is likely to progress. Genetic information plays a key role for detecting tumoral cells and tissues that can help identify cancer disease at an early stage. The role of genetic mutations in CRC has been extensively analysed and several publications are available in the literature on this topic (D. Huang et al., 2018; Oh et al., 2018; Ruiz-López et al., 2018; Valle, Vilar, Tavtigian, & Stoffel, 2019). Other authors have focused on identifying biomarkers with the aim of finding the subset with the highest predictive power (Ding, Han, Zhang, He, & Li, 2019; Kather, Halama, & Jaeger, 2018; Lech, Słotwiński, Słodkowski, & Krasnodębski, 2016; Yiu & Yiu, 2016). Early identification could increase the likelihood of survival and dramatically reduce the mortality rate. Unfortunately, a full understanding of cancer cell behaviour is still beyond our grasp, making this a major challenge in medicine.

When prevention has failed, the application of individualised therapies is the ideal scenario for the treatment of cancer patients. Personalising therapies implies finding the most suitable set of drugs and their exact dose for a given patient, based on the available input parameters, such as cancer type, tumour size, and whether metastasis is present or not. The idea behind this individualisation of therapies is to maximize the effect of drugs, limit their side effects, shorten the time necessary to cure the disease and reduce costs. The idea that individualised therapies are more cost-effective than generic ones seems credible because the same treatment is obviously not suitable for every patient since not all cases are similar. Several publications have discussed the direction that medicine is taking in this respect (Jackson & Chester, 2015; Jain, 2005; Olin, 2019; Usher-Smith et al., 2017) and its popularity has grown in recent years. Although all these authors agree that personalised treatment will increase the effectiveness of existing drugs, to the best of our knowledge, there has been no attempt to put it into practice in the case of cancer treatment, making this goal a priority in cancer research.

In this move towards individual therapies, computing sciences have become a close ally of health and life sciences and medicinal chemistry. The rapid development of high-performance computing (HPC) platforms such as parallel and distributed computing have found a place to develop in the field of chemical and biological problems. It is well known that HPC infrastructures are extensively used to carry out complex scientific calculations (Shanyu Chen et al., 2019; Schmidt & Hildebrandt, 2017; Upton, Trelles, Cornejo-García, & Perkins, 2016) and their computing power can drastically speed up the resolution of a problem (Garg, Arora, & Gupta, 2011; Nobile, Cazzaniga, Tangherloni, & Besozzi, 2017; Wang, Ma, Pratx, & Xing, 2011). However, this is not enough: firstly, because the amount of medical and pharmacological data available is overwhelming and huge computing power is needed to analyse it all; and, secondly, the analysis methods necessary to transform such data into real understandable knowledge are very challenging. While HPC can help overcome the first difficulty, the application of artificial intelligence (AI), and more specifically machine learning (ML), is necessary for the second. Only if HPC and ML work together will they be capable of screening the vast chemical space and predict the most cost-effective therapy for individual patients (Dilsizian & Siegel, 2014; Pérez-Sianes, Pérez-Sánchez, & Díaz, 2019).

Machine learning experts know that with the right data very efficient predictions can be made, as has been demonstrated in several fields such as sports results, injuries, stock market movements, text-based emotions, etc. The field of medicine has not been left behind in this respect and such technology is already used to diagnose or predict diseases such as cancer (Kourou, Exarchos, Exarchos, Karamouzis, & Fotiadis, 2015), making it clear that ML, complemented by HPC, represents the future of anti-cancer medicine. Already, ML algorithms are very helpful in many cancer-related tasks, such as the prediction and diagnosis of the disease, predicting its progression, the search for new drug synergies, predicting therapy outcomes and estimating survivability. It is the potential for analysing historical data, learning from the analysis and making predictions for future cases that makes them suitable for application in cancer research. It might even be claimed that ML is the aid that doctors need to increase the accuracy of their predictions and decision making, due to its ability to extract knowledge from previous cases. Evidently, the output of ML systems has to be transformed to make it understandable by healthcare staff; otherwise, we would be wasting an important opportunity.

This critical review highlights the role of ML in each of the main steps of anti-cancer medicine. Section 2 focuses on the needs of doctors, attempting to answer questions like "*What kind of ML do doctors need?*" and "*Does ML output need to be adapted to medical doctors?*". Section 3 presents a revision of the typical ML algorithms used in each stage, each subsection describing the most frequently used approaches, which are condensed into a table to facilitate their readability. The most relevant findings observed in Section 3 are discussed in Section 4. Finally, the main conclusions reached and the future of ML in cancer research are summarized.

## 2. What kind of ML is important in Medicine/Cancer Prediction and Treatment?

In this section, we focus on the basic features of an ML system that medical doctors and medical/biological researchers are seeking beyond the output that a trained ML system already provides.

The advantages of ML systems stem from the fact that they use thousands of features, which they use to produce decisions in a very short time. It is important to note that the training stage can be expensive in terms of computing power, while the prediction stage is in comparison fast and computationally cheaper. The correlations that the algorithm finds between the samples are similar to those found by experienced doctors, who have seen hundreds of patients and begun to notice repetitive symptoms or similar values in their detailed medical tests, which helps them to make decisions.

However, no matter how accurate ML systems are, no matter how many lives they can save in principle, and no matter if they are based on the doctor's entire medical knowledge if medical/biological researchers do not understand the underlying models and their inferences. Only if ML systems cannot be explained, these systems will not be a game changer in medicine, nor will medical/biological researchers use them to make everyday decisions, condemning the whole approach to failure. To achieve any success, ML systems need to gain trust of medical/biological researchers.

Consequently, our aim was to define four factors that should contribute to the success of ML learning systems in the medical domain: i) Output explainability, ii) Linking the predictions to the original cases used to produce outputs and iii) Low data hungriness. In this survey, we analyse existing approaches with respect to these factors as, only if there is a substantial attention paid to all of them will a novel ML approach or system be a game changer in a specific clinical situation. Only if the answer to the question "*Do doctors need to know about and learn ML in the future?*" is negative can ML add real value to clinical practice.

### 2.1. Factor One: Output Explainability

Explainability in Machine Learning or, in AI in general (XAI), is a hot topic, especially when it is applied to medicine. AI systems tend to return raw results that are hard to understand, which complicates their interpretation by non-expert users, including doctors. Thus, to make AI more attractive to healthcare professionals we should answer the question "*What do doctors need to easily interpret AI predictions?*" Explainability or interpretability often appears as a desideratum but it is poorly defined (Lipton, 2018). Hence, a clear understanding of the term explainability is essential in order to classify existing ML approaches. In general, there are two approaches to explainability: model explainability and inference explainability (Jacovi, Sar Shalom, & Goldberg, 2018). Model explainability relates to understanding how a model behaves in general, whereas inference explainability aims to describe how systems decide on each instance. Hence, these are two facets of the same problem. However, in both cases, explainability may be obtained by showing symbols (e.g., natural language or structured languages such as logical forms) to explain models or inferences.

Since the first AI systems, authors have outlined their stages of inference. For example, Swartout et al. (Swartout, Paris, & Moore, 1991) deal with explanations for expert systems, Johnson (Johnson, 1994) presents agents that learn to explain themselves, and Lacave and Diez (Lacave & Díez, 2002) discuss explanation methods for Bayesian networks (BN). In recent years, there has been a strong emphasis on revealing what happens behind the black box that uses AI algorithms (Holzinger, Langs, Denk, Zatloukal, & Müller, 2019). This is necessary if doctors are to trust the results provided by these algorithms and so use them in their daily activities (diagnosis, deciding on the most appropriate treatment, etc.). In comparison with other domains, medicine deals with the uncertain, probabilistic, unknown, incomplete, imbalanced, heterogeneous, noisy, dirty, erroneous, inaccurate and missing data sets in arbitrary high-dimensional spaces (Holzinger, Dehmer, & Jurisica, 2014; Lee & Holzinger, 2016).

Explainable artificial intelligence (XAI) has received much attention in recent years (Gunning, 2016). There are two aspects of unsupervised learning models relevant in the context of explainability (Holzinger et al., 2019). First, the representations learned in these models may show similarities between the data in a class. One such case is the word *embedding*, which can signal semantic similarity between words (Mikolov, Chen, Corrado, & Dean, 2013). Second, being able to generate instances that allow us to study the differences between data within a class. This is relevant in medicine, where the discovery and analysis of disease-related abnormalities are relevant (Schlegl, Seeböck, Waldstein, Schmidt-Erfurth, & Langs, 2017).

Trustworthiness in AI is the ability to evaluate the validity and reliability of an ML system in many different input configuration and application environments. This factor is very important in the medical environment, particularly in cancer prediction, where it is necessary to be able to evaluate exactly the limitations of an ML system and, consequently, accurately interpret and trustfully apply ML prediction system outputs.

Bærøe et al. (Bærøe, Miyata-Sturm, & Henden, 2020) underline the growing importance of AI and the relative need for trustworthiness in AI systems, especially in the medical environment. In the same work, the authors analyse the report: “Ethical guidelines for trustworthy artificial intelligence” published by the European Commission in 2019 (https://ec.europa.eu/futurium/en/ai-alliance-consultation) and highlight the need for “globalising” the guidelines at both European and international level.

### 2.2. Factor Two: Linking to original cases to produce outputs

AI systems often focus on the outputs but do not explain how much each input participates in the result. In a medical context, this correlation between inputs and outputs may be necessary to identify the reasons leading to a given decision.

Assignment methods try to link a certain output of the deep neural network with input variables (Holzinger et al., 2019). In another paper (Sundararajan, Taly, & Yan, 2017), the authors analyse the change in output gradients depending on the change in input variables. In this way, the authors propose a result based on the data that were used as the input of an algorithm and try to make a link between these data and the result obtained. However, in the medical field, although we will can still explain the results obtained and see a link with similar cases that formed the basis of a decision formulated by the AI algorithms, besides the huge opportunities that AI provides, there will always be the possibility of making a mistake and exposing the patient to certain risks (Oakden-Rayner & Palmer, 2019).

### 2.3. Factor Three: Data Hungriness

With the widespread application of computer technology in the medical field, the amount of medical data available has increased dramatically, and analysis methods are already in use for the intelligent assessment of medical health. In coming years, we expect the volume of medical data to increase even more, ranging from terabytes to petabytes and even yottabytes (Archenaa & Anita, 2015; Hermon & Williams, 2014; Ristevski & Chen, 2018).

However, due to the mixed format of medical data, incomplete records, and the noise present in them, it is still difficult to analyse large amounts of medical data (Sun, Liu, Wang, Lian, & Ma, 2019). Because traditional ML methods cannot efficiently extract a rich body of information from large medical databases, Deep Learning (DL) methods are used to build a hierarchical model by simulating the human brain. Increasingly, DL models use large medical databases, from which they select and optimize parameters and automatically learn the process of pathological analysis of doctors (Hassan, Hassan, & Kholief, 2018). Based on these models, the disease in question is in an intelligent way and an early diagnosis can be made. Thus, the pressure on the activities of doctors is considerably reduced and the efficiency of their work can be improved.

## 3. Application of ML approaches in cancer cases

In this section, a number of cases will be discussed to illustrate how ML can help doctors in the different stages of cancer evolution, from its diagnosis to the prediction of survival chances. Each section focuses on one of the main steps targeted by ML in healthcare contexts. Tables 1 to 6 summarize a detailed collection of works related with the topic of discussion. The datasets column describes the original source of data to reference a specific dataset, a full database, a citation, a project, or the institution that collected the samples. The column entitled "*Exp?*" means whether the interpretability of the results by non-experts is considered in the paper or not. Other relevant information, such as the AI approaches and the software tools used, are also reported. To facilitate the readability of the examples, we present the works in a short table per section.

### 3.1. Predict the possibility of cancer

| **Cancer type** | **AI approach** | **Datasets** | **Software** | **Training data set size** | **Data types** | **Exp?** | **Reference** |
|---|---|---|---|---|---|---|---|
| Lung | CNN | BRFSS | Caffe | 235673 | Text | Yes | (Songjing Chen & Wu, 2020) |
| Any | RF | COSMIC, dbSNP | R, HMMER, Dojo | 200, 800 | Text | No | (Kaminker, Zhang, Watanabe, & Zhang, 2007) |
| Any | SVM | Cosmic, SwissVar, Swiss-Prot | Libsvm | 6326 | Text | No | (Capriotti & Altman, 2011) |
| Breast, Thyroid, Kidney | RF | TCGA:BRCA, TCGA:THCA, TCGA:KIRP | Java, Weka, YARN, MLlib | 897, 571, 321 | Text | No | (Celli, Cumbo, & Weitschek, 2018) |
| | DT | TCGA:BRCA | unknown | 897 | Text | No | |
| | SVM | TCGA:BRCA | unknown | 897 | Text | No | |
| | BN | TCGA:BRCA | unknown | 897 | Text | No | |
| CRC | BN | NSHDS | R, Visualizations with Cytoscape | 1676 | Text | Yes | (Myte et al., 2017) |
| Breast | ANN | Private | Matlab | 62219 | Images, Text | No | (Ayer et al., 2010) |
| | CNN, SVM | unknown | R | 500 | Images, Text | No | (Heidari et al., 2018) |
| | CNN, KNN | unknown | R | 500 | Images, Text | No | |
| | GBM, SVM | KBCP, OBCS | XGBoost, Sklearn, esyN, Matplotlib, Python | 696, 923 | Text | Yes | (Behravan et al., 2018) |
| Gastric | GBM | Private | XGBoost | 1431 | Text | No | (Taninaga et al., 2019) |
| | LR | Private | unknown | 1431 | Text | No | |
| Skin | ANN | NHIS | unknown | 462630 | Text | No | (Roffman, Hart, Girardi, Ko, & Deng, 2018) |
| Ovarian | KNN, LDA, SVM, ELM | IOTA tumor images database | Matlab | 348 | Images | No | (Martínez-Más et al., 2019) |
| Cervical | CNN | Private | Caffe | 20000 | Images | No | (Martínez-Más et al., 2020) |

**Table 1.** Review of publications whose main topic is ML and cancer risk prediction.

Currently, most of the studies performed for predicting the possibility of cancer are based on the analysis of genetic data and mutations. Kaminker et al. (Kaminker et al., 2007) developed *CanPredict* software to identify and predict whether certain mutations are associated with tumours or not. The software combines the Sorting Intolerant From Tolerant (SIFT), LogR.E-value score, and Genetic Ontology Similarity Score (GOSS) methods by applying an advanced Random Forest (RF) classification scheme. Capriotti and Altman (Capriotti & Altman, 2011) used support vector machines (SVM) to analyse different databases, each created with an equal number of cancer driver Single Amino Acid Polymorphisms (SAPs) and neutral SAPs. Using this technique, it is possible to predict whether a given missense SAP is neutral or is involved in cancer appearance. In their study, the authors achieved an effectiveness greater than or equal to 90% in the overall predictions.

Taninaga (Taninaga et al., 2019) describe how a set of characteristics related to gastric cancer can be processed using extra gradient boosting decision (XGBoost) algorithms or logistic regression (LR) methods to predict whether a patient is at risk of developing the disease over the next 122 months. In this study, 10 models were developed. For the first five, the authors used XGBoost: the first model only took into account Helicobacter infections, while to the second they added data on chronic atrophic gastritis, in the third they included endoscopic findings, in the fourth they added biological background factors and in the fifth they also included blood tests. The other five models were identical applied linear logistic regression instead of XGBoost. The performance of each model was measured using the area under the curve (AUC) value. As a result of the research, the most influential characteristics in the development of gastric cancer were seen to be the mean corpuscular volume, the proportion of lymphocytes, age, body mass index (BMI), and postgastrectomy. Finally, AUC values of 0.899 and 0.874, respectively, were obtained with the 5$^{th}$ and 10$^{th}$ models, the authors concluding that with these models it is likely to predict whether a patient might suffer from cancer.

According to the American Cancer Society, 3.3 million people are diagnosed with skin cancer annually. A prediction of the risk of suffering Non-Melanoma Skin Cancer (NMSC) was made (Roffman et al., 2018) using 13 personal data of patients that can easily be obtained from an Electronic Medical Record (EMR): gender, age, BMI, diabetic status, smoking status, emphysema, asthma, race, Hispanic ethnicity, hypertension, heart diseases, vigorous exercise habits, and history of stroke. These input parameters were first normalised to values between 0 and 1 and an artificial neural network (ANN) model was developed based on one input layer with 13 nodes, two hidden layers with 13 nodes, and one output node. The authors used 462630 cases, taking 70% of the cases for training and the remaining 30% for validation, and obtained an AUC value of 0.81. The study concluded that by including the two most important factors that should be taken into account in skin cancer, i.e. radiation and personal history, risk predictions of the model could very likely be improved.

Martínez-Más et al. (Martínez-Más et al., 2019) combined different ML techniques with features obtained by Fourier transform (FT) to classify ovarian tumours as benign or malignant, using ultrasound images. After extracting 187 features from the ultrasound images using FT, they were used as input features for KNN, LD, SVM and ELM. For this, different kernels were analysed to obtain the optimal configuration, and it was seen that the combinations of FT with LD, SVM or ELM are good classifiers for biomedical images, providing an accuracy of more than 85%.

Breast cancer (BC) is one of the most common types of cancer in women. For prediction purposes, a regular analysis of mammographic images is required. To estimate the probability of malignancy of the tumour there are three categories: prognostic models, computer-aided detection, and computer-aided diagnosis. Ayer (Ayer et al., 2010) proposed a method for accurately predicting BC using ANNs, with particular emphasis on calibration made by means of the Hosmer–Lemeshow goodness-of-fit test. This generates a network topology with three layers: the first one with 36 input nodes (mammographic descriptors, demographic factors, and BI-RADS), a hidden layer with 1000 nodes and an output layer with 1 node. Later, they trained the network using a cross-validation method on 62219 registers. Next, they compared the results obtained through their model with the prediction experience of eight radiologists. The fact that the ANN obtained an AUC value of 0.965 and the radiologists a value of 0.939, demonstrates the good predictive capabilities of ANN, which can therefore be considered a reliable support tool.

Predictions of the risk of developing BC in the short term can be made by comparing the distribution of volumetric breast density of both breasts based on mammographic image analysis (Heidari et al., 2018). The authors proposed a model based on a Convolutional Neural Network (CNN), which converts an image into a characteristics vector, then applied a Locality Preserving Projection (LPP) algorithm to reduce the features obtained by the network, finally obtaining a vector with 44 characteristics. Classification was then carried out, comparing two classification methods, SVM and k-nearest neighbours (KNN). The model was trained through a cross-validation using 500 mammographic images, which provided an AUC value of 0.62 for SVM and 0.60 for KNN. In order to further optimize the accuracy of the model, the AUC values were calculated for each of the 44 characteristics and then sorted according to these values. Subsequently, the least relevant characteristics were eliminated, by testing the model based on a range of 2 to 10 characteristics. With 10 features and using KNN, an AUC value equivalent to 0.64 was obtained, which was better than when using 44 features. The best configuration was achieved using LPP-KNN, reducing the regenerated features to four. This gave an AUC value of 0.68 for the short-term prediction of BC (less than 5 years).

The risk of developing BC can be predicted through the identification of Single Nucleotide Polymorphisms (SNPs) in DNA that contribute most to its development (Behravan et al., 2018). To identify them, a three-stage protocol is implemented: (i) the SNPs are selected using a gradient boosting classification technique: XGBoost; (ii) based on the XGBoost output data, an adaptive iterative search for SNPs is made, sorting the results downwards according to their scores; the M best-scored results and the M worst-scored ones are selected and are separately ordered from lowest to highest; this process is repeated, increasing the size of M until the both lists overlap; (iii) the top SNPs are chosen and classified with SVM representing an optimal group that can potentially predict the risk of BC. The protocol is implemented in *Python* with the libraries *sklearn*, *xgboost* among others and can be downloaded from github.

DNA methylation is known to play a major role in tumorigenesis. *BIGGIOCL* (Celli et al., 2018) is a tool that can be used to analyse hundreds of thousands of individual data in a few hours. Although it was designed to analyse DNA and CpG Islands, the author specifies that it could be adapted to other fields. The tool, developed in *Java* and based on the *MLlib* learning library, allows parallelization of work in multiple machines. When developing the software one of the reasons for implementing RF was its parallelization capability that allows a forest tree to be executed in each node and the information to be sent to the master node. As it is based on *MLlib* it can be used in Yet Another Resource Negotiator (YARN) environment. In the publication the authors analysed data from HumanMethylation450 to check its relationship with BC and obtained a direct relationship with the genes RP53, PIK3CA, BRCA1, BRCA2 and BDNF, results that match those previously published by other authors.

Another type of cancer that is frequent in both men and women is CRC. Myte (Myte et al., 2017) carry out the first study relating a One-carbon metabolism (1CM) pathway to cancer risk in humans by applying a BN. The observed relationship between compounds of 1CM and CRC, and the lack of empirical studies proving the impact of 1CM and SNPs on CRC motivated this work. The study collects data from blood samples, one per patient, and uses a BN to relate population-based data, SNPs and the metabolic pathways involved in 1CM. The authors suggested that the most important factors in colorectal tumorigenesis are the associations between folate, vitamin B6 and vitamin B2, and concluded that these compounds should be taken into account in future studies of 1CM and the development of CRC.

Lifestyle is important for disease prevention. In the case of lung cancer particularly, there are certain habits or external factors that can increase the risk of contracting the disease. In the study of Chen and Wu (Songjing Chen & Wu, 2020) a set of data concerning demographics, disease, radiation, behaviour, environment, and smoking was analysed in a group of adult patients. The authors used a CNN to identify which of these factors are the most important in the development of this type of cancer. The study divided the samples into four groups: (i) men over 64 years, (ii) women over 64 years, (iii) all those over 64 years, and (iv) all those over 17 years. The four sets of data were then converted into Hierarchical Data Format 5 (HDF5), which is designed to store and organize large amounts of data and is used by *Caffe*, a Deep learning framework, to import the data into their CNNs. After training the model with a cross-validation, it achieved an AUC prediction value of 0.913 and, of all the risk factors for lung cancer examined in those over 64 years of age, smoking was the most important.

In Martínez-Mas et al. (Martínez-Más et al., 2020), the authors propose a novel method for the early detection of cervical cancer, which is one of those with high mortality in women. Frequently, the automatic classification of medical images does not pre-clean the images to remove overlaps, which does not reflect the reality of the images obtained directly from the medical samples. To overcome this issue, the authors implemented an artificial cell merger approach to improve the efficiency and realism of the classification model using CNN and without ruling out blurred, overlapping cells, etc. This approach showed a classification accuracy of 88.8%, obtaining a sensitivity and specificity of 0.92 and 0.83, respectively.

### 3.2. Predict cancer recurrence

| Cancer type | AI approach | Datasets | Software | Training data set size | Data types | Exp? | Reference |
|---|---|---|---|---|---|---|---|
| CRC | KNN, SVM, GBM, ANN, DT, RF | GEO, ArrayExpress | R | 50 | Text | Yes | (Lu et al., 2020) |
| | LR, DT, GBM | BioStudies database | Python, R | 800 | Text | Yes | (Y. Xu, Ju, Tong, Zhou, & Yang, 2020) |
| Breast | SVM, ANN, Regression | unknown | SPSS, R | 733 | Text | No | (J. Kim & Shin, 2013) |
| | SVM, ANN, DT | ICBC | Weka | 1189 | Text | No | (Ahmad, Eshlaghy, Poorebrahimi, Ebrahimi, & AR, 2013) |
| | SVM, RO | BioBIM | Java | 318 | Text | Yes | (Ferroni et al., 2019) |
| Breast | SSL | GEO, I2D | C++ | 194988 | Text | Yes | (C. Park, Ahn, Kim, & Park, 2014) |
| CRC | | | | | | | |
| Oral | BN, ANN, SVM, DT, RF | unknown | unknown | 86 | Text, Images | Yes | (Exarchos, Goletsis, & Fotiadis, 2012) |
| Cervical | SVM, DT, ELM | Chung Shan Medical University Hospital Tumor Registry | unknown | 168 | Text | Yes | (Tseng, Lu, Chang, & Chen, 2014) |

**Table 2.** Summary of studies analysed in Section 3.2 about cancer recurrence.

Once the cancer is diagnosed, one of the main concerns is the possibility of recurrence or metastasis. In this line, Exarchos et al. (Exarchos et al., 2012) used a data set comprising clinical, image and genomic data to provide a multiparametric system to detect recurrence in squamous cell carcinoma using BN, ANN, SVM, decision trees (DT) and RF classifier algorithms and ROC curve assessments. The best results were obtained for the BN classifier (78.6% for clinical data, 82.8% for images and 91.7% for genomic data). Kim et al. (W. Kim et al., 2012) studied the recurrence of BC over 5 years using SVMs, ANNs, and regression analysis; in this case, the SVM model gave the best results in terms of accuracy (89%). In the same study, it should be noted that selection of the characteristics of the models was based on the mutual information provided by the input characteristics. In the same line of detecting recurrent BC, Park et al. (C. Park et al., 2014) used genetic information to create a graphical model based on semi-supervised learning (SSL) through gene pairs that indicate strong biological interactions, in this case for both breast and colon cancer. This graphic model proved to be quite accurate in predicting the recurrence of breast and colon cancer (80.7% and 76.7%, respectively). This SSL technique was seen to very interesting when very few labelled samples are available, which is a fairly common problem for this type of data set.

In Ahmad et al. (Ahmad et al., 2013), three ML methods (DT, ANN and SVM) were compared for predicting for BC recurrence by analysing sensitivity, specificity and accuracy. The C4.5 algorithm was used in DT. Accuracy of 0.936, 0.947 and 0.957, respectively, were obtained. This work showed that SVM had the lowest error rate and the highest accuracy for predicting the recurrence of BC. In Tseng et al. (Tseng et al., 2014), SVM, DT and Extreme Learning Machine (ELM) are used to predict the recurrence of cervical cancer. Of these three methods, DT obtained the best results, especially when using the C5.0 algorithm (92.44 % accuracy). The following were analysed in the study: Pathologic Stage, Pathologic T, Cell Type and RT Target Summary.

Another way of approaching cancer prediction is through making individual predictions for each patient. Ferroni et al. (Ferroni et al., 2019) studied this approach using SVM and Random Optimization (RO) to predict BC in individual patients. In addition to prediction, the model allowed patients with low and high risk of cancer progression to be differentiated. The authors concluded that the use of ML algorithms (specifically SVM) with RO, allows the creation of an efficient model for customization in the prediction and recurrence of BC.

Two studies by Lu et al. and Xu et al. (Lu et al., 2020; Y. Xu et al., 2020) worked on the early identification of CRC recurrence. In the first paper, several treatments were analysed, and good results were observed in patients who are sensitive to FOLFOX (5-FU, leucovorin and oxaliplatin). The authors used ML algorithms (more specifically KNN, SVM, GBM, ANN, DT and RF) to identify the differences in genes between patients who respond to FOLFOX and those who do not respond in cases of CRC recurrence. They concluded that SVM and RF are the most effective ML methods for predicting FOLFOX response. In the second paper, too, ML techniques (LR, DT, Light GBM, GBM) were used to study the impact of treatments once CRC had been detected. Light GBM and GBM were found to be the most efficient for detecting the reappearance of CRC, and the treatments that most influence the reappearance of tumours were chemotherapy, age, carcinoembryonic antigen and anaesthesia time.

### 3.3. Predicting cancer progression

| Cancer type | AI approach | Datasets | Software | Training dataset size | Data types | Exp? | Reference |
|---|---|---|---|---|---|---|---|
| Lung | RF | Multicenter Clinical Trials | Matlab2016, SPSS23 | 72, 32, 31 | Images | No | (Dercle et al., 2020) |
| Lung<br>Breast<br>Renal<br>CRC | TL | TRACERx, (Yates et al., 2015), (Gerlinger et al., 2014) | ClonEvol | 768 | CCF, binary data | Yes | (Caravagna et al., 2018) |
| Lung<br>CRC | RNN | TCGA | Matlab | 506, 253 | Numbers | No | (Auslander, Wolf, & Koonin, 2019) |
| Breast | ANN | (Albertazzi et al., 1998) | unknown | 16 | Numbers | No | (Grey, Dlay, Leone, Cajone, & Sherbet, 2003) |
| Head and Neck | LR | GSE57441, GSE9844 | GraphPad Prism | 330 | Mass spectra | No | (Ishii et al., 2020) |
| Skin | Weka-FCBF, SVM, PCA, ExtraTrees, | TCGA | caret, scikit, OmicsMarkeR, Rtsne, scatterplot3d | 371, 354, 371 | Numbers | No | (Bhalla, Kaur, Dhall, & Raghava, 2019) |
| | KNN, RF, LR, Ridge | | | | | | |

**Table 3.** Works applying ML to forecast cancer progression.

Tumours can change over time, getting bigger, becoming malignant or undergoing metastasis (McGranahan & Swanton, 2017) in an evolutive process that involves cancerous cells (Greaves & Maley, 2012). Tumours evolve in different ways in different patients. The *REVOLVER* (Repeated EVOLution in cancER) method (Caravagna et al., 2018) applies the so-called Transfer Leaning (TL) approach to forecasting cancer progression. While the standard procedure infers uncorrelated models for each individual patient depicted by phylogenetic trees containing noisy data, *REVOLVER* uses TL to correlate models obtained from different patients and identify similarities in those tumours that evolve in a similar manner. The idea behind TL is to store the knowledge obtained while solving one problem and to apply this knowledge, when possible, in the resolution of a similar task. Thus, the knowledge extracted from one sample is transferred to another. As input, *REVOLVER* uses a set of Cancer Cell Fractions (CCF) or any other genetic alteration that can be represented in binary format. It then follows a two-step process: i) it calculates a set of correlated evolutionary trees, which are numerically scored, describing the evolution of each patient's tumour; and ii) it computes the evolutionary trajectories for each group of input alterations depicted in a tree that shows the number of times an alteration occurs among other values. This method was used to analyse a collection of datasets for lung, breast, renal and colorectal cancer based on 768 samples, and identified interesting genomic trajectories that were judged to merit further study (e.g. CDKNA→TP53→TERT, TP53→PIK3CA→-8p→+8q).

Alternative to TL for studying mutation timelines are Long Short-Term Memory (LSTM) networks, which are a type of recurrent neural network (RNN) with the ability to learn long-term dependencies from a sequence of events. LSTM takes advantage of the temporal nature of mutation trajectories. With this type of algorithm, mutations can be sorted by occurrence time to provide an explanation of tumour evolution (Auslander et al., 2019). The authors trained an LSTM of 5 hidden layers aiming to predict the number of mutations present in each tumour, the so-called mutational load. The model was trained on two datasets containing CRC and lung cancer samples. In less than 100 epochs they reach an AUC of 0.95. It is also possible to predict the genes that are present in such mutations and identify a set present in both types of cancer (e.g. titin, mucin-16, nesprin-1). Finally, the authors reported that the last 20 mutations are highly correlated with the mutational load. To validate their model, they implemented an SVM model that exhibited lower performance than LSTM, probably because they studied a non-linear relationship between mutations.

The state of a BC usually depends on several factors, such as the tumour size and cellularity, the presence of tumoral cells in the lymph nodes being the most reliable marker and the expression of S100A4 and nm23 genes the most effective predictors of their status. In order to investigate the predictive power of these genes and tumour size and grade a set of 15 ANNs was trained on 16 BC samples and tested against another 16 (Grey et al., 2003). The results confirmed the expression of S100A4 and nm23 genes as the most effective predictor and that the inclusion of other markers could improve the accuracy (e.g. ER/PgR expression).

Simpler ML approaches, such as LR, can also help in predicting cancer progression (Ishii et al., 2020). The method works in the knowledge that Transforming Growth Factor beta (TGF-β) is involved in the acquisition of heterogeneity by tumours (Hall & Massagué, 2008). This fact means that TGF-β is responsible for promoting tumour evolution, thus complicating cancer prognosis. The activation of TGF-β signalling contributes to the acquisition of malignant properties by head and neck squamous cell carcinoma (HNSCC). However, the effects of TGF-β on lipid metabolism remain unclear. In this context, the authors aimed to develop an ML-based algorithm to detect intratumoral TGF-β-stimulated areas in clinical HNSCC tissue without recourse to a conventional immunohistological examination. For this purpose, Logistic Regression of the mass spectra of HNSCC-stimulated and non-stimulated human cells was carried out on the public datasets GSE57441 and GSE9844. The LR algorithm accurately segregated stimulated and non-stimulated cells reaching a classification accuracy of up to 98%. This finding demonstrates that simple ML approaches, despite their limitations, can also be helpful in predicting cancer progression.

Metastatic Skin Cutaneous Melanoma (SKCM) has been demonstrated to arise from factors such as the expression of mRNAs and miRNAs and aberrations in methylation patterns (Greenberg, Chong, Huynh, Tanaka, & Hoon, 2012; Mazar et al., 2011). To understand how skin melanoma progresses a combination of feature selection methods and ML classifiers has been used (Bhalla et al., 2019). The data, including mRNA, miRNA and methylation expressions from The Cancer Genome Atlas (TCGA) database, were split into 80% for training and 20% for testing, giving training datasets of 371, 354 and 371 samples respectively. First, three feature selection methods, namely Weka-FCBF, SVM with L1 regularization (SVM-L1) and Principal Component Analysis (PCA), were applied to reduce the number of input features so that subsequent analysis could focus on the most discriminative characteristics. In this step, SVM-L1 outperformed the other methods by selecting the 17 features that were used in the next stage. The Jaccard index was calculated to select the best method. Secondly, six classification models were developed to support vector classification with weight (SVC-W) performed best, obtaining 0.95 AUC and 89.4% accuracy in an external validation test. The other classifiers were ExtraTrees, KNN, RF, LR and Ridge classifier. The models were assessed using different metrics, including AUC, the Matthews coefficient, sensitivity, specificity and accuracy. As a conclusion, the authors reported a collection of genes that could be considered relevant markers of cutaneous melanoma metastasis (e.g. ESM1, NFATC3, C7orf4).

### 3.4. Calculating drug doses or drug combinations

It used to be commonly accepted that the administration of drug combinations rather than providing monotherapy can increase treatment efficacy (Mokhtari et al., 2017). This approach is nowadays limited by the huge size of the chemical space that makes the identification of novel drugs very difficult and, consequently, complicates the choice of effective drug combinations. In order to perform a cost-effective screening of this chemical space, DL methods are gaining in importance. For example, the *DeepSynergy* tool (Preuer et al., 2018) aims to predict the most efficacious anti-cancer multi-drug treatments by means of DL. *DeepSynergy* provides an ANN, which is implemented with the modern *TensorFlow* framework, and outperformed other ML methods, such as gradient boosting machines (GBM), RF, SVM and Elastic Nets, in a benchmark on the largest synergy dataset. However, the performance all these methods decreased when exploring new datasets of different sizes and data distributions, which is one of the typical problems of ML approaches which remains a challenge today. In the same line, Celebi (Celebi, Bear Don't Walk, Movva, Alpsoy, & Dumontier, 2019) published a study to identify functional anti-cancer dual therapies, an approach whereby two single-target drugs work in synergy to cure a disease. The above authors evaluated five ML methods (LR, Lasso, SVM, RF and GBM) implemented with the *sklearn* and *xgboost Python* libraries. All the models were trained on a novel dataset released by AstraZeneca and the Dialogue for Reverse Engineering Assessments and Methods consortium (Menden et al., 2017). The assessment showed that GBM outperformed the other methods in synergy identification. It is interesting to mention that the study included a variant of LR, the so-called Lasso (Tibshirani, 1996), which is a regularized version of LR that reduces overfitting in the model.

| **Cancer type** | **AI approach** | **Datasets** | **Software** | **Training dataset size** | **Data types** | **Exp?** | **Reference** |
|---|---|---|---|---|---|---|---|
| Prostate | ANN | UCSD #140520 study | unknown | 66 | Text, Images | unknown | (Shiraishi, Tan, Olsen, & Moore, 2015) |
| | ANN | UCSD #140520 study | unknown | 66 | Text, Images | No | (Shiraishi & Moore, 2016) |
| | CNN | unknown | Keras, Tensorflow | 72 | Images | No | (Nguyen et al., 2019) |
| Breast | DSS | Local database | unknown | unknown | DB-stored medical records | Yes | (Musen, Tu, Das, & Shahar, 1996) |
| Any | LR, SVM, RF, GBM | AstraZeneca, DREAM consortium | sklearn, xgboost | 2790 | Numbers | Yes | (Celebi et al., 2019) |
| | MVA on Undirected Graphs | GDSC, CCLE, CTRP | R, Matplotlib, Graphviz | 700 | CSV, Text | Yes | (Keshava et al., 2019) |
| | ANN | (O’Neil et al., 2016) | TensorFlow | 23062 | Compounds, Cell lines | Yes | (Preuer et al., 2018) |
| | RF | Princess Margaret Cancer Centre | unknown | 383 | Images | No | (McIntosh & Purdie, 2016) |
| | CNN | PASCAL VOC 2012 | TensorFlow | 1464 | Images | No | (L.-C. Chen, Papandreou, Kokkinos, Murphy, & Yuille, 2017) |
| | CNN | PASCAL VOC 2012 | Caffe, TensorFlow | 1464 | Images | No | (L.-C. Chen, Papandreou, Schroff, & Adam, 2017) |
| | ANN | NCI database | unknown | 141 | Text | Yes | (Weinstein et al., 1992) |

**Table 4.** Manuscripts applying ML to estimate drug doses or finding drug combinations for cancer therapies.

Besides deciding on the drug combination to be administered, identifying the exact dose is crucial for creating personalised cancer therapies. However, despite the importance of these points, research into them lags behind estimating cancer risk or predicting therapy outcome. *EON* software (Musen et al., 1996), a component-based decision support system (DSS) that was developed to build healthcare protocols at a high level of abstraction, represented a first attempt to use AI to build reusable software capable of helping doctors. Its modular design makes it easy to add and replace components, and the graphical interface means that it is accessible to any user, even those lacking advanced computer skills. A major advantage of *EON* is that, once designed, the protocols can be reused for any disease with minimal adaptations; for example, different types of cancer or AIDS might share the same protocol. With regards to drug dose estimation and the optimal application time, *EON* includes the *Chronus* temporal query system, which implements a specific algebra for writing temporal queries and can be extended with the *Catenation* operator. This operator is able to identify adjacent periods and merge them into a single one, making it possible to know when and for how long a patient was given a certain drug combination. This information, along with the therapy outcomes for the same periods, can help analyse the effectiveness of a drug synergy, providing useful information for future cases.

A recently published work (Kearney, Chan, Valdes, Solberg, & Yom, 2018) summarizes the main advances of AI for treating head and neck cancer patients. A key factor when planning treatments for this cancer is the intensity modulated radiotherapy (IMRT) dose prediction. The manuscript describes the way ANN (Shiraishi & Moore, 2016; Shiraishi et al., 2015), CNN (L.-C. Chen, Papandreou, Kokkinos, et al., 2017; L.-C. Chen, Papandreou, Schroff, et al., 2017; Nguyen et al., 2019) and tree-based methods (McIntosh & Purdie, 2016) are currently applied to resolve classification problems from a collection of images. The aim of this sort of protocol is to identify the most effective dose for each patient. Tree-based methods try to mimic the thinking of an expert clinician looking at a set of images of a new patient, identify a similar past patient with the most similar images, and map the dose distribution administered to the former patient in order to assess the optimal treatment to be applied with the new patient. To do this, a collection of features is extracted from the images to build a dataset of structured data that can be handled by most ML algorithms. This approach reached 78.68% and 86.83% accuracy in breast and prostate cancer, respectively, when the Gamma metric was used. The main drawback of tree-based algorithms that work in this way is that their accuracy is closely coupled to their core steps: extracting descriptive features from the source images, identifying a similar patient on the basis of such descriptive features and adapting the past dose to the new patient. The alternatives to the tree-based methods used in the above work are fully connected ANNs with two layers, which are easy to train but which do not conserve memory and may suffer overfitting. Whatever the case, the prediction error reported was lower than 10% (Shiraishi & Moore, 2016). Fortunately, CNNs are very good for predicting volumetric information, the most suitable types being Tiramisu and Dilated CNNs (DCNN). Tiramisu models work in two steps: i) encoding the input image to extract the most descriptive features; and ii) decoding the information to restore it to the initial size. When the dose volumes are consistent with respect to the anatomy (e.g. in prostate cancer),

Tiramisu models are the preferred option (Nguyen et al., 2019), otherwise (e.g. head and neck cancer), DCNNs are preferable.

Frequently, gene mutations are detected in cancer patients, and discovering the relationship between these genetic variations and drug responses has led to the ability to identify which patients might profit most from certain drug synergies. However, the results of clinical trials in their advanced stages must exhibit a significant improvement over standard therapy. Thus, clearly defining groups of patients in which a novel drug may be more effective than the existing ones could help lead to individualised therapies, and, as a consequence, this has become a target of ML. An unsupervised learning approach based on multivariate analysis (MVA) of undirected graphs (Keshava et al., 2019) was performed to classify patients into well-defined subpopulations. The statistical methods were implemented with *R* packages and the input datasets were collected from the GDSC (https://www.cancerrxgene.org/), CCLE (https://portals.broadinstitute.org/ccle) and CTRP (https://portals.broadinstitute.org/ctrp/) databases. As result of this work, the SEABED (Segmentation and Biomarker Enrichment of Differential Treatment Response) platform was developed and used in several examples, in one of which the authors aimed to assess the response to a combination of drugs, namely A and B. To accomplish this, they segmented patients into subpopulations depending on their response to the therapies, considering AUC and $IC_{50}$ as metrics. They also provided a graphical representation of the results in a tree whereby the identified subpopulations were coloured depending on the exhibited sensitivity to both, A, B or no drugs, which is important for facilitating interpretation of the results. Then, the authors chose a BRAF and a MEK inhibitor and discovered that the subpopulation sensitive to A was enriched for BRAF mutations and the one sensitive to B was enriched for MEK mutations. This approach is generic enough to be used for the analysis of any type of cancer sample, independently of its particular characteristics and can also be of great use for predicting tumour progress.

As can be inferred from Table 4, image processing is a key procedure when estimating drug doses and finding effective drug combinations. To satisfy the need for powerful image processing algorithms, CNNs have shown themselves to be alternative to traditional ANNs. In parallel, new frameworks (e.g. *TensorFlow*, *PyTorch*) have been developed to exploit all the computing power of graphical processing units (GPUs) and accelerate image analysis. When there are no images available or their inspection is not suitable, other statistical methods and classifiers (e.g. LR, RF, MVA) can be fed with a diverse collection of data types. Regarding interpretability of the results, this is not the main concern of scientists according to Table 4. Very few of the works try to adapt the output of their models to make it understandable by doctors or use easily interpretable models (e.g. DT, BN). Whatever the case, the extensive use of image processing with CNN makes some models easier to understand than raw numerical results.

### 3.5. Predict treatment outcome

| Cancer type | AI approach | Datasets | Software | Training dataset size | Data types | Exp? | Reference |
|---|---|---|---|---|---|---|---|
| CRC | CNN | Akershus University Hospital, Aker University Hospital, Gloucester Colorectal Cancer Study, VICTOR trial | TensorFlow | $12*10^6$ | Images | No | (Skrede et al., 2020) |
| | RF | Teikyo University Hospital, Gifo University Hospital | unknown | 54 | Medical Records | No | (Tsuji et al., 2012) |
| | RF, SVM, ANN, DT, KNN, GBM | GSE19860, GSE28702, GSE72970 | caret, class, e1071, gbm, tree, randomForest, RSNNS | 50 | Raw data | No | (Lu et al., 2020) |
| | LR, DT, GBM | BioStudies database | Scikit-learn, R | 800 | Excel | No | (Y. Xu et al., 2020) |
| | BN | ACTUR database | NCSS | 5301 | DB-stored | Yes | (Steele et al., 2014) |
| | | | | | medical records | | |
| | RF, ANN | Genomics of Drug Sensitivity in Cancer portal | Encog, randomForest | 38930 | Raw data | No | (Menden et al., 2013) |
| | SVM | GSE19860, GSE28702, GSE72970 | e1071 | 144 | Raw data | No | (Lin, Qiu, Zhang, & Zhang, 2018) |
| | RF | GSE52735, GSE62080, GSE69657 | limma, glmnet, Boruta, randomForest, pROC | 58 | Raw data | No | (Gan et al., 2019) |
| | SVM, LR | unknown | Orange | 38 | unknown | No | (Land, Margolis, Gottlieb, Yang, & Krupinski, 2010) |
| | SVM | Val d'Aurelle Regional Cancer Center | MAS 5.0 | 5 to 19 | Numbers | No | (Del Rio et al., 2007) |
| Breast | Diagonal LDA, KNN | Nellie B. Connally Breast Center, M.D. Anderson Cancer Center, Instituto Nacional de Enfermedades | dCHIP | 133 | Text, Numbers | No | (Hess et al., 2006) |
| | | Neoplásicas de Lima | | | | | |
| | SVM, Recursive Feature Elimination | University of Heidelberg | e1071, ROC | 52, 48 | Numbers | No | (Thuerigen et al., 2006) |
| | LR | unknown | unknown | 84 | Numbers | No | (Harris et al., 2007) |
| Bladder | DT | University of Southern California | SPSS | 948 | Numbers | No | (Mitra, Skinner, Miranda, & Daneshmand, 2013) |
| Blood | LDA | FRALLE93 protocol | unknown | 32 | Numbers | No | (Talby et al., 2006) |
| Renal | SVM | National Wilms Tumor Study-5 | e1071 | 250 | Numbers | No | (C. C. Huang et al., 2009) |
| Ovary | Binary LR, Stochastic Regression | Duke University Medical Center, H. Lee Moffitt Cancer Center and Research Institute | Bioconductor | 83 | Numbers | No | (Dressman et al., 2007) |
| Esophageal | SVM | unknown | unknown | 46 | Text, Numbers | No | (Duong et al., 2007) |
| Lung | DT, RF, ANN, SVM, LR, GBM | (Belderbos et al., 2005), (Bots et al., 2017), (Carvalho et al., 2016), (Janssens et al., 2012), (Jochems et al., 2017), (Kwint et al., 2012), (Lustberg et al., 2016), Morin (forthcoming), (Oberije et al., 2015), (Olling, Nyeng, & Wee, 2018), (Wijsman et al., 2015), (Wijsman et al., 2017) | caret | 156, 137, 363, 179, 327, 139, 922, 257, 548, 131, 149, 188 | Text | Yes | (Deist et al., 2018) |
| Head and Neck | | | | | | | |
| Meningioma | | | | | | | |
| Laryngeal | | | | | | | |

**Table 5.** List of works presented in Section 3.5 about the prediction of therapy outcome in cancer patients.

In the move towards personalised therapies, the prediction of therapy outcome is essential. In spite of the fact that several works where AI is used to estimate a tumour's evolution after therapy for colorectal (Del Rio et al., 2007; Tsuji et al., 2012), breast (Harris et al., 2007; Hess et al., 2006; Thuerigen et al., 2006), blood (Talby et al., 2006), renal (C. C. Huang et al., 2009), ovary (Dressman et al., 2007) or oesophageal (Duong et al., 2007) cancer, this topic remains a major challenge for scientists.

Classification, regression and clustering algorithms have frequently been used to resolve this sort of issue. As example of the classification method, a DT was implemented to diagnose and predict therapy outcome for bladder cancer patients using the *SPSS* statistical package (Mitra et al., 2013). The work showed how nearly 950 patients could be classified into three groups with different recurrence-free and overall survival probabilities. DTs have the advantage of being very intuitive and easy to interpret by medical doctors, which is one of the main aims of health-related MLs. A similar statistical analysis for classification purposes was carried out with BN implemented with *Number Cruncher Statistical Systems (NCSS)* on a dataset of CRC patients (Steele et al., 2014). In this case, the positive prediction rate ranged from 78 to 84 per cent when estimating recurrence for the training dataset extracted from the ACTUR database. The main limitation of this work is data reliability and consistency due the military nature of some institutions feeding the data source, which lack approved programs for cancer treatments. RF is another widely used recurrent classification algorithm that is already used to predict the response to FOLFOX (5-FU, leucovorin and oxaliplatin) therapy (Tsuji et al., 2012). The model was able to correctly predict 69.2% of cases in the test set. Relationships between genomic alterations and drug responses is a factor that could lead to enhanced individual therapies. Although both genomic features and chemical properties have been computationally analysed, there is still a lack of works studying both factors together. To shed some light on this topic, ANNs and RF were used to predict therapy outcomes (Menden et al., 2013). The core of this work was the implementation of a three-layer ANN. The inputs were 608 cell lines and 111 drugs, a number between 1 and 30 hidden nodes were tested to find the best performing architecture and the $IC_{50}$ predicted value was the only output. Note that the $IC_{50}$ value is normalized in the range [0,1] by the sigmoid function added in the output layer. Based on the $R^2$ performance metric, the model obtained 0.64 on the test dataset extracted from the GDSC portal, and 0.61 on an external validation dataset. Then, a RF implemented in *R* was developed to ascertain whether the ANN model could be improved but it resulted in a $R^2$ of 0.59 on the blind test dataset, which is a slightly lower value than that achieved by the ANN model. Although the results look promising, the model has some limitations that could be overcome by adding more cell lines, epigenetics data and gene expression data as inputs. Classification algorithms could also help in identifying potential biomarkers too which is another topic that has received increasing attention in recent years. An *R*-implemented RF (Gan et al., 2019) for this task achieved 81% accuracy in the validation dataset. A feature selection step is carried out in this study before the classification. Reducing the dimension of the input makes the classifier faster and facilitates interpretation of the results by clinicians.

The diversity of classification and regression algorithms makes scientists wonder about the best choice to build new models and benchmark their own. To fairly assess some of the most typical classifiers an extensive study was carried out (Deist et al., 2018) with a set of algorithms. Six classifiers were evaluated on twelve datasets related to different cancers (lung, head, neck, meningioma, and laryngeal) using the AUC as a measure of which ones will work well in the future too. Although none of the algorithms stood out over the others, RF and Elastic Net Logistic Regression (ENLR) exhibited a higher discriminative power in chemo and radiotherapy outcome. Therefore, it is suggested that they might be the first choice when building classification models. The authors also claim that RF and ENLR should be the preferred option against which custom models should be compared.

Many other supervised learning approaches can be found in the literature. Most of the cases exploit datasets from the National Center for Biotechnology Information (NCBI) or collected from local institutions. SVMs represent a method that is commonly adopted to predict tumour progress after therapy and is especially helpful when predicting FOLFOX therapy results in CRC patients because this type of algorithm usually works with images. When working alone, SVM reached a positive prediction rate of 85.4% (Lin et al., 2018), which is similar to that obtained by RF. But SVM can also be combined with LR to provide a novel scoring method to measure the tumour size response to therapy, as it outperforms the traditional WHO and RECIST measurements (Land et al., 2010).

Recent studies assessed a variety of ML methods in CRC prediction scenarios. Lu (Lu et al., 2020) compared six models implemented with *R* packages in a FOLFOX response prediction task. The models represented the following approaches: RF, SVM, ANN, DT, KNN, and GBM. The experimental tests showed that RF and SVM were the most accurate methods when predicting FOLFOX outcome. Unfortunately, their performance fell off when predicting other therapies such as FOLFIRI (5-FU, leucovorin, and irinotecan), therefore their application to future patients is limited. The reason for this reduction in performance when using alternative therapies seems to be related with the aforementioned use of unexplored datasets with different characteristics, which would indicate a close relationship between the model and the training data. The third best-ranked classifier was the ANN model, whose accuracy was close to that of RF and SVM but was more consistent when confronted with other therapies. This result demonstrates that ANNs constitute a powerful predictive tool for future CRC studies. In another work (Y. Xu et al., 2020) the authors assessed four ML methods (LR, DT, GBM, and Light GBM) and found GBM and Light GBM to be more accurate than the others. This evidence leads us to think that GBM probably gain in importance in the near future. Finally, the rapid development of ANN and its variants (e.g. recurrent neural networks, convolutional neural networks, adversarial neural networks) has encouraged scientists to develop enhanced and more powerful networks capable of profiting from HPC architectures. As a result of that evolution, several libraries (e.g. *TensorFlow*) are widely used nowadays. Tensor-based networks are especially useful for image processing due to their ability to exploit all the computing power of GPUs to analyse images in a parallel manner. This novel ML paradigm has been used to build a CNN model that anticipates the outcome after resection based on a dataset of 12 million images (Skrede et al., 2020).

The poor interpretability of the results is a challenge that needs to be faced. Raw estimations or complicated charts might be unintelligible to doctors and may render any ML algorithm worthless for practical reasons. The data types feeding ML systems intended to predict therapy outcomes are very different, ranging from binary data to well-structured records (e.g. Excel, CSV, database records). In this step, the application of image processing through CNN is not so frequent, as explained in the previous section, but still constitutes the preferred approach when manipulating images, as can be seen in Table 5.

### 3.6. Predicting likelihood of survival

Once cancer has been diagnosed, classified, and treated, the next questions are how the tumour will evolve and how likely is the patient's survival. The former was already answered in section 3.3, so this section will focus on the available ML methods for the latter. Note that the works introduced in section 3.5 not necessarily predict the survival chances in months, for example, but is more likely to focus on how the treatment will reduce the tumour size. The prognostication of a patient's survivability is not easy and depends on many factors, such as the type of cancer and the stage. Fortunately, ML can help doctors evaluate survival chances by analysing several biomarkers in a systematic manner. With the aim of answering this question, Zhu (Zhu, Xie, Han, & Guo, 2020) summarizes an extensive collection of works concerning the use of DL in cancer prognosis, including some that estimate the survival likelihood and even the survival time.

| Cancer type | AI approach | Datasets | Software | Training dataset size | Data types | Exp? | Reference |
|---|---|---|---|---|---|---|---|
| Breast | SVM | (van de Vijver et al., 2002) | unknown | 295 | Numbers | No | (X. Xu, Zhang, Zou, Wang, & Li, 2012) |
| | BN | (Van't Veer et al., 2002) | unknown | 97 | Numbers | Yes | (Gevaert, De Smet, Timmerman, Moreau, & De Moor, 2006) |
| | SSL | SEER database | unknown | 162500 | DB-stored medical records | No | (K. Park et al., 2013) |
| | SSL Co-training | SEER database | unknown | 162500 | DB-stored medical records | No | (J. Kim & Shin, 2013) |
| | ANN, LR, DT | SEER database | unknown | 200000 | DB-stored medical records | Yes | (Delen, Walker, & Kadam, 2005) |
| Oral | SVM | unknown | unknown | 69 | unknown | No | (Rosado, Lequerica-Fernandez, Villallain, Peña, et al., 2013) |
| Any | ANN | unknown | unknown | 440 | unknown | No | (Y.-C. Chen, Ke, & Chiu, 2014) |
| Lung | Linear Regression, DT, SVM, GBM, Custom[a] | SEER database | R | 7830 | DB-stored medical records | Yes | (Lynch et al., 2017) |
| CRC | CNN, RNN | Helsinki University Central Hospital | Keras | 420 | Images | Yes | (Bychkov et al., 2018) |
| Brain | CNN | TCGA, South Australian public hospital system | Keras, Tensorflow | 679 | Images | Yes | (Zadeh Shirazi et al., 2020) |
| Prostate | DT, BN, Cox | The Methodist Hospital | S-PLUS | 1050 | Text | Yes | (Zupan, Demšar, Kattan, Beck, & Bratko, 2000) |

**Table 6.** Summary of works about ML and the likelihood of survival

[a] A custom ensemble of methods.

According to a recent review (Kourou et al., 2015), SVMs provide the most accurate predictions of cancer survival. Although all the analysed studies are trained on small datasets, they are able to reach up to 98% and 97% accuracy in oral (Rosado, Lequerica-Fernandez, Villallain, Pena, et al., 2013) and breast (X. Xu et al., 2012) cancer, respectively. Other approaches such as ANNs (Tseng et al., 2014) and BNs (Gevaert et al., 2006) are showing good results as well, attaining more than 83% accuracy, and both are expected to gain in importance in coming years. On the other hand, SSL, which only works with a few labelled samples, has emerged as a feasible alternative to the classic supervised and unsupervised learning paradigms but, as its results show (71% and 76% accuracy reported by (K. Park et al., 2013), and (J. Kim & Shin, 2013)), predictive capacity of this approach still has to be improved. Nevertheless, another study on lung cancer that used similar ML techniques yielded different results (Lynch et al., 2017). The authors evaluated linear regression, DT, SVM, GBM, and a custom ensemble, finding that GBM was the most accurate model in terms of root mean square error (RMSE). All the models were implemented in *R* language and trained on SEER database. In recent decades, cancer has been one of the preferred fields for the assessment of ML models to predict survival likelihood. An analysis of survivability in prostate cancer patients (Zupan et al., 2000) was carried out using three non-linear statistical methods: DT, BN and Cox (Cox, 1972). This work represents a case study that aims to demonstrate that ML classifiers are useful for estimating a patient's survival chances, a process that is receiving increasing attention from ML experts. The authors conclude that ML statistical models could be helpful in the near future for predicting survival and other issues such as the probability of recurrence in cancer patients.

The new wave of ML is dominated by ANNs and their subtypes such as convolutional, recurrent or adversarial neural networks, among others. CRC can also profit from ANNs to predict survival chances, especially when the input datasets are image collections and the use of CNNs is advantageous. A recent work (Bychkov et al., 2018) described the training of a DL system, built on convolutional and recurrent neural networks, to classify tumour images. Such classifications of tumour images are a frequent way of predicting tumour evolution and, consequently, evaluating survival chances. It is worth mentioning that the classifier used by these authors ran on a GPU to accelerate the processing and deliver the results in a short time. GPUs can speed up CNN calculations dramatically, which is a huge advantage due to the large number of samples that CNNs usually deal with and the high number of layers they have. Other cancer types also take advantage of CNNs and exploit GPU computing power. Such is the case with brain cancer, for which condition patient survival can be estimated by means of the recently published classifier *DeepSurvNet* (Zadeh Shirazi et al., 2020). *DeepSurvNet* builds CNN models implemented with *Keras* and *TensorFlow* libraries, which are trained with a dataset from the TCGA Program (Weinstein et al., 2013). The models classify the patients into four groups, each with an estimated overall survival.

The use of ML approaches whose output can be graphically represented, such as BN, DT and CNN, facilitates the interpretation of survival chances by healthcare professionals. The easy interpretation of results should always be taken as a requirement when ML is to be applied in a context outside computer sciences. It is also worth noting that medical records extracted from public databases are a common input (Hutter & Zenklusen, 2018; SEER, 2020) when evaluating survivability, which indicates that long-term well-structured data are the most useful data source to predict survival chances.

## 4. Software and datasets

In this section, we will summarize the most relevant technical details extracted from the above-mentioned works, such as the software tools created, the availability of the source code, the use of HPC platforms and the main features of the datasets. Figure 2 summarises the approaches applied at every stage. It can be observed that ANN, LR and SVM are the most common methods in cancer research. RF, BN, DT, KNN, GBM and CNN are also used frequently but are not reported in all the tasks.

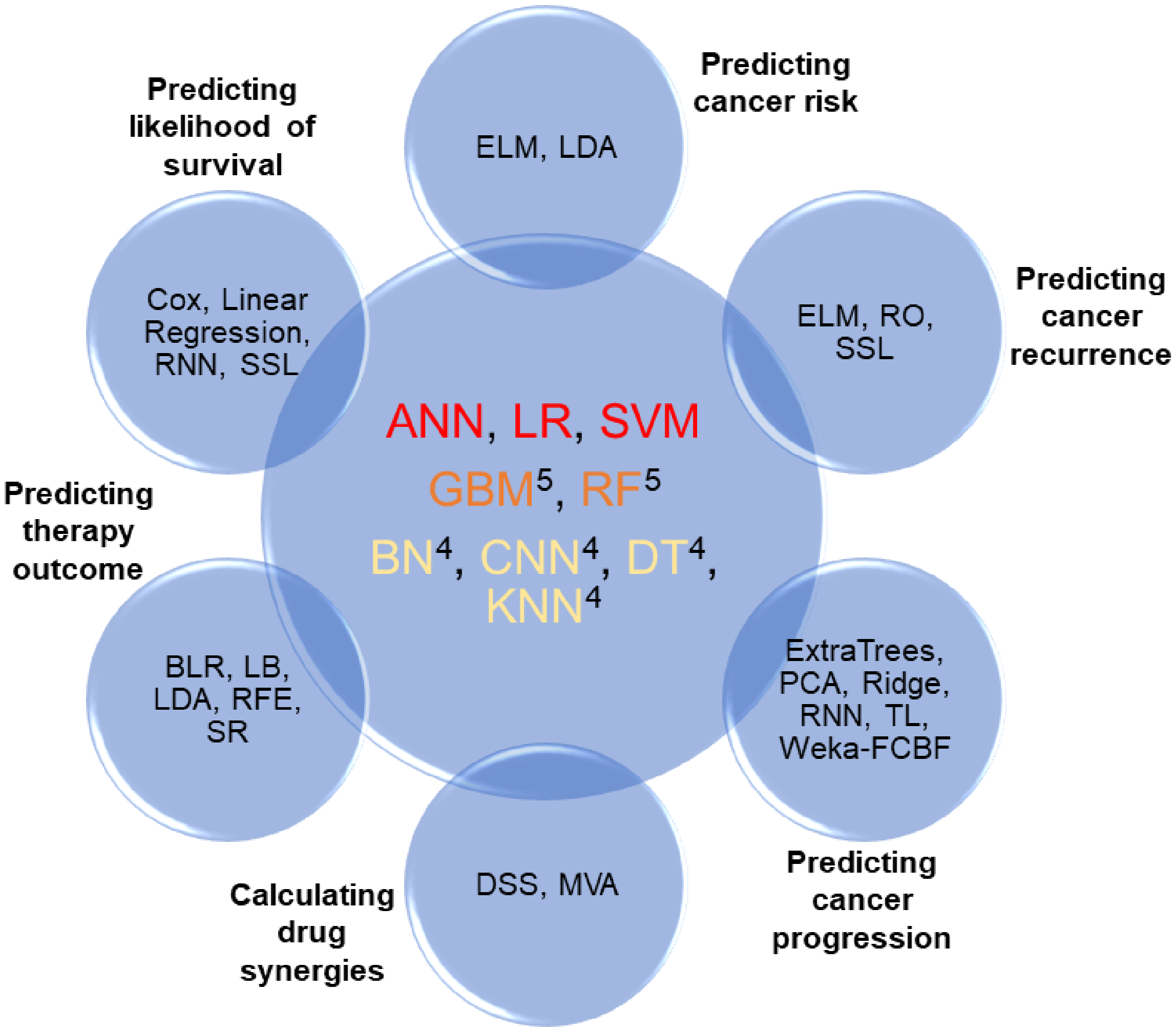

**Fig. 2.** Graphical summary of ML methods being applied in cancer research tasks. Super indices in the central figure represent the number of steps in which that approach is reported. No index means that the approach is reported in all the tasks.

### 4.1. Software tools

In Tables 1-6 we have enumerated a number of libraries and frameworks frequently used for developing ML models. There is a clear trend to implement models in *R* and *Python* languages. *R* is a good choice for rapid development due to the diverse collection of packages it provides (e.g. *caret*, *e1071*, *Bioconductor*) and the many possibilities it offers to create different models, including SVM, RF and DT, among others. Therefore, its simplicity and flexibility make it an attractive alternative for several scientists. The other preferred option is *Python* and, in particular, frameworks like *TensorFlow* and *PyTorch*.

Tensor-based frameworks have gained in importance in recent years supported by the rapid development of GPUs, which are a very suitable hardware solution for tensored calculations. While the use of *R* implementations has been mentioned for several years, publications reporting works in *Python*-based frameworks tend to date from 2017. This confirms the intuition that the development of GPUs, and more generally HPC, will be closely connected with the advances achieved in the performance of ML algorithms in the near future.

Many statistical tools are less frequently used. This group of statistical methods is composed of tools such as *Matlab*, *SPSS*, *Caffe* and *Weka*. Although they are not so powerful as programming languages, they offer many statistical features that allow the rapid development of models, including LR, SVM, ANN and BN. Furthermore, the indicated tools are well established in the academic world, and so many scientists are familiar with them and their reliability has been extensively proved.

Despite being well known and a very stable language, *Java* is barely used in this context. Only the *Encog* and *MLlib* libraries are reported in the works. There may be many reasons to explain this but the main ones are probably that *Java* is usually considered slower than other languages and that the users do not have the programming skills required by this tool.

| **Task** | **Reference** | **Code availability** |
|---|---|---|
| Predict cancer risk | (Behravan et al., 2018) | https://github.com/hambeh/breast-cancer-risk-prediction |
| | (Celli et al., 2018) | https://github.com/fcproj/BIGBIOCL |
| Predict progression | (Auslander et al., 2019) | https://github.com/noamaus/LSTM-Mutational-series |
| | (Bhalla et al., 2019) | https://webs.iiitd.edu.in/raghava/cancerspp/ |
| Predict recurrence | (W. Kim et al., 2012) | http://ami.ajou.ac.kr/bcr/ |
| Estimate drug synergy | (Musen et al., 1996) | https://protege.stanford.edu/ |
| | (Celebi et al., 2019) | https://github.com/rcelebi/dream-drugcombo https://www.synapse.org/#!Synapse:syn5605365/wiki/394725 |
| | (Keshava et al., 2019) | https://github.com/szen95/SEABED |
| | (Preuer et al., 2018) | http://www.bioinf.jku.at/software/DeepSynergy/ |
| | (L.-C. Chen, Papandreou, Kokkinos, et al., 2017) | https://github.com/tensorflow/models/tree/master/research/deeplab |
| | (L.-C. Chen, Papandreou, Schroff, et al., 2017) | http://liangchiehchen.com/projects/DeepLab.html |
| Predict therapy outcome | (Deist et al., 2018) | https://github.com/timodeist/classifier_selection_code |
| Predict survival | (K. Park et al., 2013) | http://embio.yonsei.ac.kr̃/Park/ssl.php |

**Table 7.** List of code repositories or servers listed in the manuscript.

Few authors share the source code of their models with the community (see Table 7). Sometimes they prefer to develop and release a novel tool providing the obtained models through a web interface (Bhalla et al., 2019; Preuer et al., 2018). While this is an understandable decision it hinders understanding of the models by external users. However, other researchers freely share their codes, usually on github, and allow others to study and analyse how they are developed. From an objective point of view, this is the preferred solution because it allows existing codes to be better understood, improved and optimized, as well as the development of new models from a solid base.

### 4.2. HPC infrastructures

While HPC platforms are rarely reported in the analysed papers, the use of GPUs has increasingly been mentioned recent years (e.g. Bychkov et al., 2018; Zadeh Shirazi et al., 2020). The recent development of Tensor-based frameworks and libraries for ML, e.g. *TensorFlow*, *Keras*, *PyTorch*, has promoted the use of GPUs for programming ML algorithms (Bychkov et al., 2018; L.-C. Chen, Papandreou, Kokkinos, et al., 2017; Y.-C. Chen et al., 2014; Nguyen et al., 2019; Preuer et al., 2018; Skrede et al., 2020; Y. Xu et al., 2020; Zadeh Shirazi et al., 2020). The rapid integration of GPU computing in ML strongly suggests that faster ML algorithms will emerge in coming years, resulting in the ability to handle even larger training datasets.

Please note that, although references to other HPC paradigms have not been found in this revision, it is very possible that other authors have leveraged HPC platforms (e.g. parallel computing) in their works.

### 4.3. Datasets

We can broadly classify the input datasets into two major groups: i) those obtained from publicly available databases; and ii) those collected from institutions (e.g. hospitals or universities). Although both online and custom approaches are valid, public datasets facilitate the reproducibility of the experiments. SEER and TCGA databases are typically used in cancer research.

Leaving their source aside, we have focused on two properties of the datasets: the data types they contain and the size of the training dataset. The data types vary widely between works, including in terms of the text, images, medical records and binary data. Numerical values are the preferred option for feeding ML algorithms because they mostly work on numerical calculations. As can be observed in Tables 1-6, when public or private institutions are responsible for collecting data, they usually work with numerical data. Also, text inputs are widely used, probably because they can be easily translated into numerical values. Images are typically used to feed CNNs due to the ability of this type of network to apply sequential filters on images and extract patterns. This is also a frequent option because many hospitals and universities have easy access to historical images from scans, tomography or mammography.

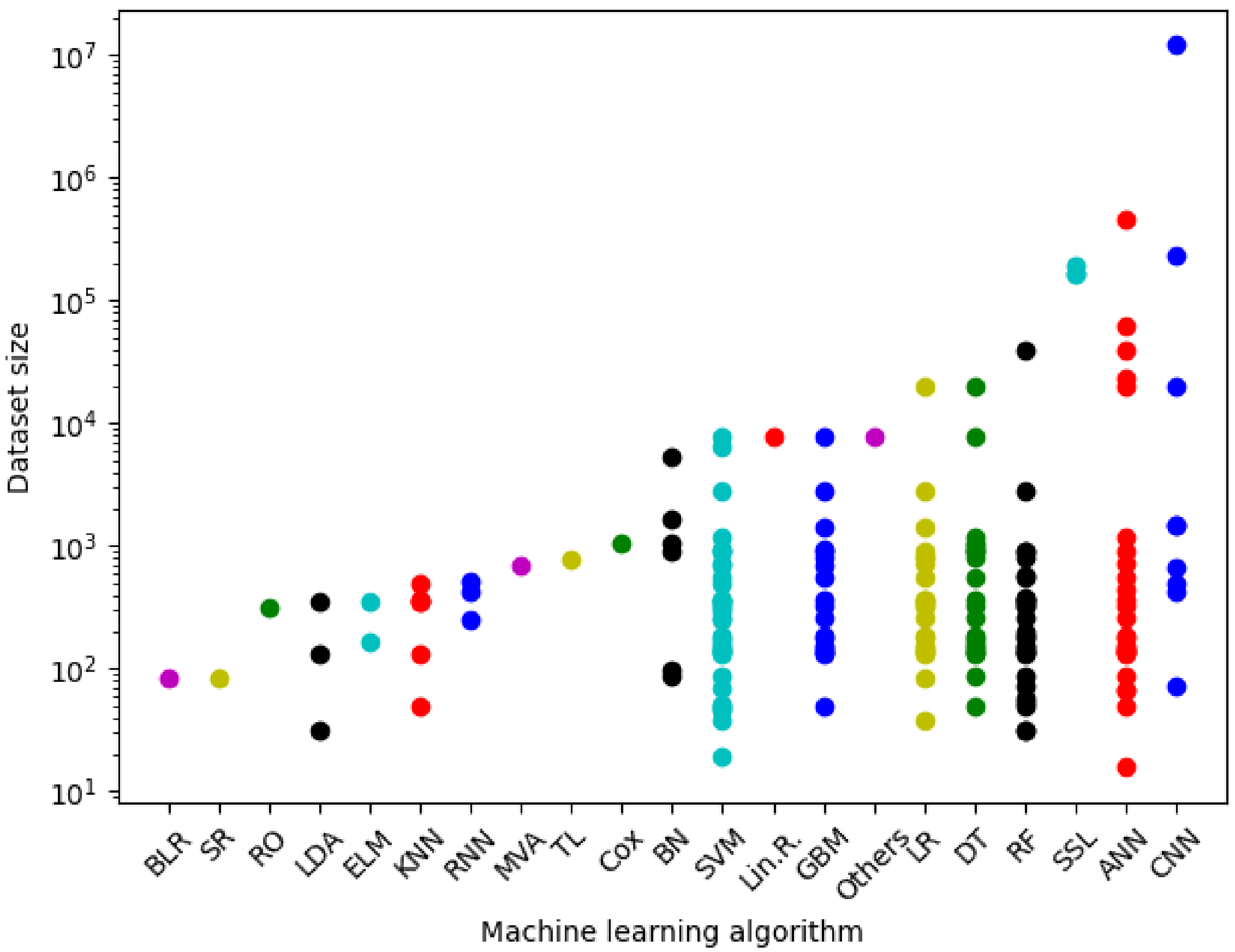

**Fig. 3.** Reported dataset size by algorithm.

Structured information is more suitable for ML than unstructured. Usually, ML algorithms receive a set of well-defined inputs, which they evaluate and weigh to make predictions. Therefore, when the inputs are clearly defined, the models can be easily developed (e.g. BN, ANN, LR, DT). This is the case of databases, such as ACTUR, SEER and TCGA, and other datasets where the fields are undoubtedly separated.

The second key feature of the analysed datasets is their size, which ranges from tens to millions of samples. In general terms, neural networks (ANN, CNN, RNN) handle the largest datasets (e.g. 200000, 463080, 235673 and $12 \times 10^6$ samples). Although a large number of samples may seem an advantage, their sheer numbers can slow the system down during training. Thus, finding a good balance between dataset size and learning capability is required. By contrast, the simplest approaches seem to need fewer data to learn as can be inferred from the fact that the smallest datasets (less than 100 training samples) are used by traditional methods such as RF and SVM. Figure 3 shows the reported dataset size used in ML algorithms.

## 5. Conclusions and outlook

Decision-making is one of the main challenges in modern medicine is exercised at every stage of a disease's lifecycle, from diagnosis to the prediction of recurrence. Traditionally, doctors have trusted their experience to choose the best option for individual patients. However, they cannot be expected to recall all the details of all the patients they have treated in the past, which clouds their ability to recognise patterns in similar situations. This is where computational help is required.

In recent years, AI, and, more specifically, machine and deep learning, have looked at medical decision-making. In this context, anti-cancer medicine has been found to be a favourite playground due to the high mortality rate of the disease, the increasing number of cases expected in the forthcoming years and the vast amount of data available in databases of hospitals, universities and research centres. The diversity of existing cancer types encourages experiments with different ML algorithms aimed at the same target. In this review, we have analysed the generalised use of ML in cancer research but always bearing in mind CRC.

CRC is the fourth cause of mortality due to cancer worldwide and the number of cases that are expected to appear in the next decade is not promising, making it a suitable target for ML. Any ML algorithm can be applied on CRC research ranging from the simplest (e.g. LR, SVM, KNN) to the most complex ones (e.g. CNN, DNN) but it has been observed that ANN, LR and SVM are frequently reported in any task related with decision-making (risk prediction, recurrence prediction, tumour progression, estimation of drug synergy, therapy outcomes and survival time estimates). Moreover, RF, GBM, BN, DT, KNN and CNN are often applied in many cases.

There is no clear relationship between the selected approach and the type of data feeding the system. However, CNN is clearly the preferred option when manipulating medical images. It is clear that well-structured records with text or numerical fields are the simplest and favourite options when available. The dataset size is another key factor when training ML or DL systems. If the dataset is too small, the ML system will face difficulties related with learning and generalizing, whereas excessively large datasets may slow down the training phase. Thus, finding the optimal dataset size remains a challenge. As regards performance in terms of computing time, his key concern for scientists has resulted in the emergence of libraries and frameworks specifically focused on profiting from HPC facilities, such as GPUs. GPU are the preferred architecture for running CNN calculations, and NVIDIA has placed its bet on this technology becoming the world's leading manufacturer.

Explainability has been identified as the third key point to worry about, although it is no less important than the typical accuracy and performance metrics. The importance of explainability stems from the fact that ML is increasingly used in a medical context, where users are often inexperienced in interpreting AI metrics and results. Consequently, output must be translated into a language that physicians can understand. It has been perceived that explainability is still barely considered in most of the works analysed, suggesting that it is a factor that can be improved in order to "democratise" AI in many other areas. To improve the explainability of systems, feature selection methods are sometimes applied before classification. This technique helps to reduce the input size leading to faster classification and providing a more interpretable output. Some ML algorithms such as BN and DT are especially appropriate for this purpose because they return labelled directed graphs which are very easy to read and interpret.

In short, we predict a bright future of ML and DL in medical decision-making but the results must be more explainable in this or any other context. Identifying the optimal training dataset size is another factor that deserves further study. Fortunately, the rapid development of HPC will make ML systems more efficient and enable them to transform the overwhelming quantity of historical data stored in public and private databases into real, reliable and valuable knowledge.

**Funding**

This work has been funded by grants from the European Project Horizon 2020 SC1-BHC-02-2019 [REVERT, ID:848098]; Fundación Séneca del Centro de Coordinación de la Investigación de la Región de Murcia [Project 20988/PI/18]; and Spanish Ministry of Economy and Competitiveness [CTQ2017-87974-R].

**Competing interests**

The authors declare that they have no competing interests.

**Author's contributions**

All the authors contributed equally to this work.

**Acknowledgments.**

This work has been funded by grants from the European Project Horizon 2020 SC1-BHC-02-2019 [REVERT, ID:848098]; Fundación Séneca del Centro de Coordinación de la Investigación de la Región de Murcia [Project 20988/PI/18]; and Spanish Ministry of Economy and Competitiveness [CTQ2017-87974-R]. Powered@NLHPC: This research was partially supported by the supercomputing infrastructure of the NLHPC (ECM-02).